\documentclass[letterpaper]{article} 
\usepackage{aaai2026}  
\usepackage{times}  
\usepackage{helvet}  
\usepackage{courier}  
\usepackage[hyphens]{url}  
\usepackage{graphicx} 
\usepackage{natbib}  
\usepackage{caption} 
\usepackage{algorithm}
\usepackage{algorithmic}

\usepackage{newfloat}
\usepackage{listings}
\DeclareCaptionStyle{ruled}{labelfont=normalfont,labelsep=colon,strut=off} 
\floatstyle{ruled}
\newfloat{listing}{tb}{lst}{}
\floatname{listing}{Listing}
\usepackage{amsmath}
\usepackage{amssymb}
\usepackage{multirow}
\usepackage{booktabs,subcaption}
\usepackage{pifont}
\usepackage{dsfont}

\title{Retrieving Objects from 3D Scenes with \\ Box-Guided Open-Vocabulary Instance Segmentation}

\author {
    Khanh Nguyen, 
    Dasith de Silva Edirimuni, 
    Ghulam Mubashar Hassan, 
    Ajmal Mian
}
\affiliations {
    The University of Western Australia\\
    duykhanh.nguyen@research.uwa.edu.au,\\
    \{dasith.desilva, ghulam.hassan, ajmal.mian\}@uwa.edu.au
}

\begin{document}

\maketitle

\begin{abstract}
Locating and retrieving objects from scene-level point clouds is a challenging problem with broad applications in robotics and augmented reality. This task is commonly formulated as open-vocabulary 3D instance segmentation. Although recent methods demonstrate strong performance, they depend heavily on SAM and CLIP to generate and classify 3D instance masks from images accompanying the point cloud, leading to substantial computational overhead and slow processing that limit their deployment in real-world settings. Open-YOLO 3D alleviates this issue by using a real-time 2D detector to classify class-agnostic masks produced directly from the point cloud by a pretrained 3D segmenter, eliminating the need for SAM and CLIP and significantly reducing inference time. However, Open-YOLO 3D often fails to generalize to object categories that appear infrequently in the 3D training data. In this paper, we propose a method that generates 3D instance masks for novel objects from RGB images guided by a 2D open-vocabulary detector. Our approach inherits the 2D detector’s ability to recognize novel objects while maintaining efficient classification, enabling fast and accurate retrieval of rare instances from open-ended text queries. Our code will be made available at \url{https://github.com/ndkhanh360/BoxOVIS}.
\end{abstract}
\section{Introduction}
\label{sec:intro}
In recent years, advances in deep learning have been a key driving force behind the rapid progress of artificial intelligence, enabling a wide range of real-world applications across diverse domains \cite{latent_diffusion,glamor,ldlva,trellis,ct_scangaze,xray_eyemovement}. However, significant challenges remain in complex real-world settings, particularly when intelligent systems such as service robots and autonomous agents are required to localize and retrieve objects of interest in 3D environments based on open-ended user queries.
Addressing these challenges demands models that can effectively connect natural language with 3D spatial understanding, extending information retrieval beyond text or images into the 3D physical world. The goal of open-vocabulary 3D instance segmentation (OV-3DIS) is precisely this: given a 3D scene point cloud with multi-view RGB-D images and a text query, the model needs to identify and output the 3D masks of relevant objects, regardless of whether their categories were seen during training. This capability bridges multimodal retrieval and spatial reasoning, enabling more natural and context-aware interactions with 3D environments.

Most prior work on 3D instance segmentation (3DIS) \cite{mask3d,isbnet} focuses on closed-set recognition, where models are trained to segment and label objects from a limited, predefined taxonomy. Although effective for known classes, such models fail to generalize to novel objects, significantly limiting their usability in real-world applications. To overcome this, recent OV-3DIS methods \cite{openmask3d,open3dis,any3dis,openyolo3d} leverage pre-trained 2D vision-language models such as CLIP \cite{clip} and segment-anything model (SAM) \cite{sam} to extend recognition ability to unseen categories. Typically, these approaches classify 3D instance masks using CLIP embeddings; the masks themselves are produced either by pre-trained 3DIS models \cite{mask3d,isbnet} or by lifting 2D segmentations from SAM \cite{grounded_sam,sam2} into 3D. However, these pipelines are computationally expensive, requiring multiple passes through large 2D foundation models and often taking 5-10 minutes per scene. This high latency makes existing OV-3DIS systems impractical for dynamic retrieval tasks in interactive or time-sensitive settings.

To reduce the running time, Open-YOLO 3D \cite{openyolo3d} was recently introduced as a fast open-vocabulary 3DIS framework that removes the dependency on SAM and CLIP. By combining Mask3D \cite{mask3d}, a pre-trained class-agnostic 3DIS model, for mask generation with a lightweight open-vocabulary 2D detector YOLO-World \cite{yoloworld} for classification, it achieves an inference speed of around 22 seconds per scene while maintaining strong recognition performance. However, Open-YOLO 3D relies entirely on the pretrained 3D segmenter to produce candidate objects, which often misses instances of rare classes that are poorly represented in existing 3D datasets. Following the efficiency-focused direction of Open-YOLO 3D, we propose a box-guided proposal generation method to construct 3D masks for novel objects based on YOLO-World’s predicted 2D boxes. Unlike existing works, our method does not depend on SAM to create instance masks; instead, it leverages superpoints from an efficient graph-based segmentation algorithm \cite{superpoints_seg} to assemble the 3D instances. This design allows us to retain the efficiency of Open-YOLO 3D while inheriting the strong generalization ability of the 2D foundation detector. Our approach consistently achieves higher performance compared to Open-YOLO 3D while maintaining processing time under one minute per scene. Furthermore, our method can retrieve low-frequency instances from 3D scenes where Open-YOLO 3D often fails to detect. In summary, our contributions are as follows:
\begin{enumerate}
    \item We propose a method that leverages 2D bounding boxes to guide the discovery of novel object masks in 3D point clouds, improving generalization to unseen categories.
    \item We introduce a mask generation strategy that avoids the computationally heavy SAM, reducing computation and enabling inference in under one minute per scene.
    \item We demonstrate that our approach improves performance on rare objects across two popular benchmarks while maintaining high efficiency, highlighting the effectiveness of the proposed method.
\end{enumerate}

\begin{figure*}[h!]
    \centering
    \includegraphics[width=0.9\linewidth]{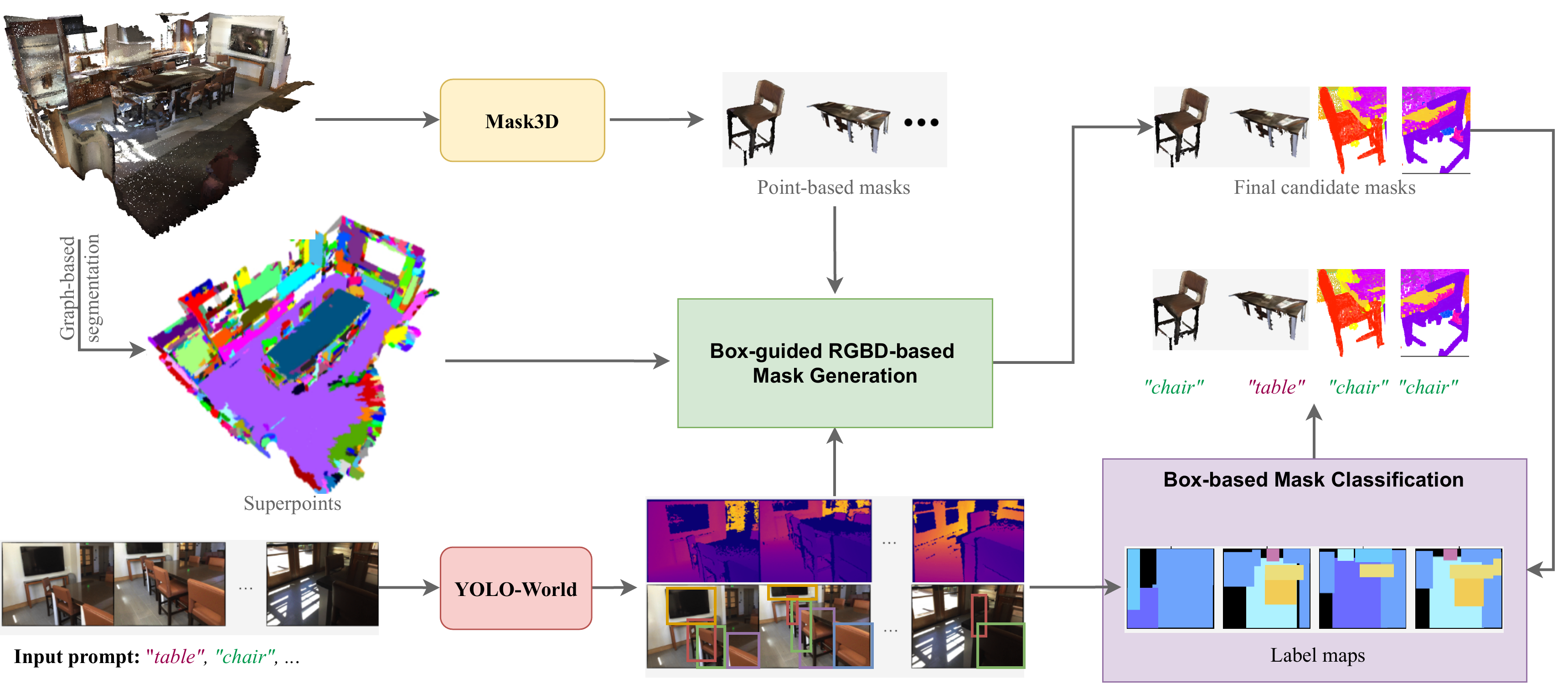}
    \caption{Overview of the proposed method.}
    \label{fig:overview}
\end{figure*}
\section{Related Work}
\label{sec:related_work}
\paragraph{Open-Vocabulary 2D Scene Understanding} aims to recognize both classes seen during training as well as novel classes at test time. Given a text query containing class names, existing methods can locate and output either bounding boxes \cite{yoloworld,grounding_dino,lp_ovod,detic,detclipv2} or instance masks \cite{sam,sam2,grounded_sam,seem,openseed} for relevant objects in an image. To transfer recognition ability from seen to unseen categories, most approaches leverage the pretrained visual-text embeddings of CLIP \cite{clip} or ALIGN \cite{align} as joint image-text representations that encode both base and novel classes. Trained on large-scale image-text corpora, these models demonstrate strong generalization to unseen categories and have therefore been adopted by recent methods for open-vocabulary recognition in 3D scenes. In our work, we use YOLO-World \cite{yoloworld}, a real-time 2D object detector, to assist in recognizing novel objects within 3D point cloud scenes.

\paragraph{Open-Vocabulary 3D Instance Segmentation (OV-3DIS).}
Inspired by advancements in 2D vision, several OV-3DIS methods have emerged to retrieve objects in 3D point clouds based on user-defined queries. Given a prompt consisting of class names, these methods predict a set of 3D instance masks and assign each mask a class label corresponding to the queried categories. Most approaches adopt a two-stage pipeline: (1) generating 3D mask proposals, and (2) classifying each proposal into a class or prompt ID.

\noindent For the first stage, some works \cite{openmask3d,openyolo3d} directly use the outputs of class-agnostic pretrained 3D segmenters \cite{mask3d,isbnet} to obtain point-based 3D masks. Others \cite{ovir3d,sai3d} generate 3D masks solely from RGB-D images by leveraging powerful open-world 2D detection and segmentation models \cite{grounding_dino,sam,detic,grounded_sam,sam2}. Point-based masks are often more geometrically accurate, yet 3D segmenters struggle to generalize to rare or unseen categories due to limited 3D training data. To address this, several methods \cite{open3dis,any3dis} fuse both point-based and RGBD-based proposals. However, RGBD-based generation typically requires running both a 2D detector and a segmentation model, resulting in significant computational overhead and slow inference.

\noindent For the second stage, most OV-3DIS approaches extract CLIP \cite{clip} features from projected image regions corresponding to each 3D mask across multiple views, and then match these features to the input prompt based on feature similarity. This multi-view feature extraction is computationally expensive and significantly increases the runtime. Open-YOLO 3D \cite{openyolo3d} alleviates this bottleneck by using 2D detector predictions directly for classification, substantially accelerating inference.

\noindent In this paper, we leverage a 2D object detector to guide RGBD-based mask generation, without relying on any 2D segmentation model, and to classify 3D instance candidates. This design preserves computational efficiency while achieving strong generalization to novel categories.
\section{Method}
\label{sec:Method}
Fig.~\ref{fig:overview} illustrates our method.  
The input to the pipeline consists of a 3D point cloud $P$ of the scene with $N$ points, accompanied by $K_{f}$ RGB images $\mathcal{I} = \{\mathcal{I}_i \in \mathbb{R}^{3\times W \times H} \mid \forall i \in (1, ..., K_{f})\}$ as well as depth information from multiple views, with camera intrinsics $I \in \mathbb{R}^{N_f\times 4\times4}$ and extrinsics $E \in \mathbb{R}^{N_f\times 4\times4}$ matrices. The task is to retrieve and localize objects in a 3D point cloud that match a given text query by generating their corresponding 3D instance masks.  

\noindent First, the point cloud $P$ is processed using a graph-based segmentation method \cite{superpoints_seg} to obtain geometrically coherent regions (called superpoints), which serve as the 3D primitives for forming novel instances. 
We employ the pretrained 3D instance segmenter Mask3D \cite{mask3d} to extract binary instance masks from $P$ (referred to as point-based masks), denoted as $M^{point}\in \mathbb{Z}_2^{K_{point}\times N}$, where $K_{point}$ is the number of point-based masks and $\mathbb{Z}_2 = \{0, 1\}$. 

\noindent From the RGB images, we use YOLO-World \cite{yoloworld}, a 2D open-vocabulary object detection model, to generate bounding boxes for relevant objects in each frame $\mathcal{I}_i$. These boxes, along with their semantic information, are used to construct label maps $\mathcal{L}_i$, which are later used to assign semantic labels to each 3D mask proposal.  

\noindent Since 3D networks are prone to missing rare objects, we leverage depth and camera information to uplift 2D boxes to 3D and gradually form novel 3D masks from superpoints (Sec.~\ref{subsec:box_proposal}). These RGBD-based masks are then combined with the point-based masks to obtain the final set of 3D instance candidates. Finally, each 3D proposal is projected into the 2D label maps to obtain a prompt distribution, which is used to match the instances to the input query (Sec.~\ref{subsec:instance_cls}).

\subsection{Box-Guided RGBD-Based Mask Generation}
\label{subsec:box_proposal}
For every RGB image $\mathcal{I}_i$, we generate a set of $K_{bi}$ bounding boxes 
$\text{B}_i = \{ (b_{ij}, c_{ij}) \mid b_{ij} \in \mathbb{R}^4, \hspace{0.1cm} c_{ij}\in \mathbb{N}, \forall j \in (1, ..., K_{bi}) \}$ 
using an open-vocabulary 2D object detector, where $b_{ij}$ are the bounding box coordinates and $c_{ij}$ is its predicted class label.  
Given a 2D bounding box $b_{ij}$, we project all pixels within the box to 3D using the corresponding depth information and camera parameters. We then leverage Open3D \cite{open3d} to obtain a 3D oriented box $b_{ij}^{3D}$ that contains all the projected points. To avoid redundancy, we filter out boxes that overlap with existing point-based masks from the 3D segmenter. Specifically, a 3D box $b_{ij}^{\mathrm{3D}}$ is considered redundant and removed if it overlaps with at least $\tau_{\mathrm{box}}\%$ of the points in a point-based mask, indicating that both correspond to the same object.

For the remaining boxes, we extract the superpoints contained within each box. A superpoint is assigned to a box if at least $\tau_{\mathrm{spp}}\%$ of its points fall inside the box. The resulting superpoint set forms a coarse mask for box $b_{ij}^{\mathrm{3D}}$, denoted as $S_{ij}$. We then merge these coarse masks sequentially across frames to construct the RGBD-based object candidate set $S = \{ S_j \mid \forall j \in (1, \ldots, K_{\mathrm{RGBD}}) \}$, where $S_j$ represents the superpoint set of the $j$-th object candidate and $K_{\mathrm{RGBD}}$ denotes the total number of RGBD-based masks.

We start by adding all superpoint sets from the first frame to $S$. In each subsequent frame $i$, each coarse mask $S_{ij}$ is compared with all masks in $S$: if it has an IoU of at least $\tau_{merge}$ and shares the same prompt label with an existing candidate $S_k \in S$, it is merged with $S_k$ to represent the same object; otherwise, it is treated as a new candidate and added to $S$. Merging is performed by combining the superpoints in $S_k$ with the new superpoints in $S_{ij}$. This process continues until all coarse masks have been merged or added to the candidate set. The candidate set is then converted to binary masks, called RGBD-based masks and denoted as $M^{RGBD}\in \mathbb{Z}_2^{K_{RGBD}\times N}$ with $\mathbb{Z}_2 = \{0, 1\}$.

We perform an additional filtering step to remove novel masks that largely overlap with point-based proposals. Specifically, a novel mask $M^{\mathrm{RGBD}}_i$ is discarded if its IoU with any point-based mask $M^{\mathrm{point}}_j$ exceeds $\tau_{\mathrm{filter}}$. This design prioritizes rare objects missed by the class-agnostic 3D segmenter; when overlap occurs, we retain the point-based mask, which typically provides higher-quality geometry. Finally, the RGBD-based and point-based proposals are merged into a final set $M \in \mathbb{Z}_2^{K \times N}$, where $K$ is the total number of instances.

\subsection{Box-Based Mask Classification}
\label{subsec:instance_cls}
The results of the 2D detector are not only used to guide the discovery of novel objects but also to match the 3D instances  with the correct classes in the input query. As shown in previous work \cite{openyolo3d}, this approach eliminates the need for computationally heavy CLIP \cite{clip}, significantly improving the efficiency of the pipeline. We follow the approach proposed by Open-YOLO 3D, which consists of the following steps:

\subsubsection{Constructing Label Maps.}
For every RGB image $\mathcal{I}_i$ with predicted bounding boxes
$\text{B}_i = \{ (b_{ij}, c_{ij}) \mid b_{ij} \in \mathbb{R}^4, \hspace{0.1cm} c_{ij}\in \mathbb{N}, \forall j \in (1, ..., K_{bi}) \}$, we construct a label map $\mathcal{L}_i \in \mathbb{Z}^{W \times H}$ by initializing all elements to $-1$, representing no relevant class label. We then sequentially replace the elements inside a bounding box $b_{ij}$ with its corresponding predicted prompt label $c_{ij}$, processing the bounding boxes from largest to smallest. The intuition is that when two objects of different sizes are visible from the same camera viewpoint, the smaller object is only seen if it is closer to the camera than the larger one; therefore, the corresponding area should be assigned the label of the smaller object.

\subsubsection{Computing Visibility.}
We first project the point cloud $P$ onto all frames to obtain $P^{2D} \in \mathbb{R}^{N_f\times 4 \times N}$ in a single shot as follows: $ P^{2D} = (I \star E) \cdot P$, where $\star$ is batch-matrix multiplication and $\cdot$ is matrix multiplication.

\noindent Then, we compute the visibility $V^f\in \mathbb{Z}_2^{N_f\times N}$ of the projected points within all frames as:
\begin{equation*}
    V^f = \mathds{1}(0<P_{x}^{2D} < W)\odot \mathds{1}(0<P_{y}^{2D} < H),
\end{equation*}
where $P_x^{2D} \in \mathbb{R}^{N_f \times N}$ and $P_y^{2D} \in \mathbb{R}^{N_f \times N}$ are the $x$ and $y$ coordinates of the projected 3D points on all $N_f$ frames.

\noindent Additionally, the occlusion visibility matrix $V^d\in \mathbb{Z}_2^{N_f\times N}$ is calculated as
$V^d = \mathds{1}(|P_z^{2D} - D_z| < \tau_{depth})$,
where $D_z \in \mathbb{R}^{N_f\times N}$ is the depth value of the point cloud obtained from the corresponding depth maps, and $P_z^{2D} \in \mathbb{R}^{N_f \times N}$ is the depth value of the projected points.

\subsubsection{Aggregating Class Distribution.}
For a 3D mask candidate $M_j$, its prompt distribution is defined as:
\begin{equation*}
\mathcal{D}_j = \{ \mathcal{L}_i[P_{i, x}^{2D}\cdot  M_{ji}, P_{i, y}^{2D}\cdot M_{ji}] \mid \forall i \in \mathcal{P}_k \},
\end{equation*}
where $M_{ji} = V_i^d \cdot V_i^f \cdot M_j \in \mathbb{Z}_2^N$ is the visibility for the $j$-th 3D mask in the $i$-th frame, $\mathcal{P}_k$ is the set of top $k$ frame indices where the $j$-th 3D mask has the most visibility, and $[\cdot, \cdot]: \mathbb{Z}^{W\times H} \mapsto \mathbb{Z}^{n}$ is a coordinate-based selection operator with $n$ being an arbitrary natural number.  

\noindent Intuitively, $\mathcal{D}_j$ is the aggregated set of class labels of all visible projected pixels of the $j$-th object in the top visible frames. The class distribution is computed as the occurrence of each class ID $c$ in $\mathcal{D}_j$, and the instance is assigned to the category with the highest probability. For more details, please refer to \cite{openyolo3d}.
\section{Experiments}
\label{sec:experiments}
\subsection{Experimental Setup}
\begin{table*}[h!]
\centering
\Huge
    \setlength\aboverulesep{0pt}\setlength\belowrulesep{0pt}
    \setlength{\tabcolsep}{6.5pt}  
    \resizebox{\textwidth}{!}{
    \begin{tabular}{@{}cccccccccccccc@{}}
    \toprule
    \multirow{2}{*}{Method} &
      \multicolumn{2}{c}{3D proposals} &
      \multirow{2}{*}{\begin{tabular}[c]{@{}c@{}}2D \\ detector\end{tabular}} &
      \multicolumn{2}{c}{SAM} &
      \multirow{2}{*}{CLIP} &
      \multirow{2}{*}{mAP} &
      \multirow{2}{*}{mAP$_{50}$} &   
      \multirow{2}{*}{mAP$_{25}$} &   
      \multirow{2}{*}{mAP$_h$} &      
      \multirow{2}{*}{mAP$_c$} &      
      \multirow{2}{*}{mAP$_t$} &      
      \multirow{2}{*}{time/scene (s)} \\ \cline{2-3} \cline{5-6}
     &
      \rule{0pt}{3.5ex}\begin{tabular}[c]{@{}c@{}}Point-\\ based\end{tabular} &
      \begin{tabular}[c]{@{}c@{}}RGBD-\\ based\end{tabular} &
       &
      \begin{tabular}[c]{@{}c@{}}Proposal \\ generation\end{tabular} &
      \begin{tabular}[c]{@{}c@{}}Mask \\ labeling\end{tabular} &
       &
       &
       &
       &
       &
       &
       &
       \\
       \midrule
       \midrule
    \rule{0pt}{2ex}Mask3D (supervised)       & $\checkmark$ &        &        &        &        &        & 26.9 & 36.2 & 41.4 & 39.8 & 21.7 & 17.9 & 13.41  \\
    \midrule
   \rule{0pt}{2ex} OVIR-3D       &        & $\checkmark$ & $\checkmark$ & $\checkmark$ &        & $\checkmark$ & 13.0 & 24.9 & 32.3 & 14.4 & 12.7 & 11.7 & 466.8  \\
    OpenMask3D    & $\checkmark$ &        &        &        & $\checkmark$ & $\checkmark$ & 15.4 & 19.9 & 23.1 & 17.1 & 14.1 & 14.9 & 553.87 \\
    Open3DIS      & $\checkmark$ & $\checkmark$ & $\checkmark$ & $\checkmark$ &        & $\checkmark$ & 23.7 & 29.4 & 32.8 & 27.8 & 21.2 & 21.8 & 360.12 \\    
    Any3DIS       & $\checkmark$ & $\checkmark$ & $\checkmark$ & $\checkmark$ &        & $\checkmark$ & 25.8 & -    & -    & 27.4 & 23.8 & 26.4 & N/A    \\
    \midrule
    \rule{0pt}{2ex}Open-YOLO3D   & $\checkmark$ &        & $\checkmark$ &        &        &        & 24.7 & 31.7 & 36.2 & 27.8 & 24.3 & 21.6 & 21.8   \\
    Ours          & $\checkmark$ & $\checkmark$ & $\checkmark$ &        &        &        &  24.9 (+0.2) & 32.1 (+0.4) & 36.8 (+0.6) & 27.6 (-0.2) & 24.3 (+0.0) & 22.4 (+0.8) & 55.9 \\
    \bottomrule
    \end{tabular}
    }
\caption{Performance and runtime comparison on the ScanNet200 validation set.}
\label{tab:scannet200}    
\end{table*}
\paragraph{Datasets and Evaluation Metric.}
We conduct experiments on the ScanNet200 \cite{scannet200} and Replica \cite{replica} datasets. ScanNet200 consists of 1,201 training and 312 validation scenes collected from diverse indoor environments. It covers 198 object categories, grouped into head, common, and tail subsets based on category frequency. Following prior works \cite{openmask3d,open3dis,openyolo3d,detailmatters}, we report results on the validation set.  
We also evaluate on Replica \cite{replica}, a synthetic dataset containing digital replicas of 8 real-world indoor scenes, covering 48 object classes. The input prompts for the models are the lists of object categories defined by each dataset.  
For evaluation, we use mean average precision (mAP) computed over IoU thresholds from 50\% to 95\% in 5\% increments as the primary metric. We additionally report mAP at IoU thresholds of 50\% and 25\%, denoted as mAP$_{50}$ and mAP$_{20}$, respectively.

\paragraph{Implementation Details.}
We use the pretrained Mask3D \cite{mask3d} to generate class-agnostic point-based masks, following prior OV-3DIS methods \cite{openyolo3d,open3dis,openmask3d}. To construct label maps, we run the YOLO-World \cite{yoloworld} extra-large variant on the first frame of every 10-frame interval for ScanNet200, and on all frames for Replica, ensuring a fair comparison with Open-YOLO 3D.  
For efficiency, we downsample each processed frame by a factor of 5 during the box-guided RGBD-based proposal generation stage.

\subsection{Quantitative Results}
\begin{table*}[ht]
\centering
\Huge 
    \setlength\aboverulesep{0pt}\setlength\belowrulesep{0pt}
    \setlength{\tabcolsep}{6.5pt}  
    \resizebox{0.8\textwidth}{!}{%
    \begin{tabular}{ccccccccccc}
    \toprule
    \multirow{2}{*}{Method} &
      \multicolumn{2}{c}{3D proposals} &
      \multirow{2}{*}{\begin{tabular}[c]{@{}c@{}}2D \\ detector\end{tabular}} &
      \multicolumn{2}{c}{SAM} &
      \multirow{2}{*}{CLIP} &
      \multirow{2}{*}{mAP} &
      \multirow{2}{*}{mAP$_{50}$} &   
      \multirow{2}{*}{mAP$_{25}$} &   
      \multirow{2}{*}{time/scene (s)} \\ \cline{2-3} \cline{5-6}
     &
      \rule{0pt}{3.5ex}\begin{tabular}[c]{@{}c@{}}Point-\\ based\end{tabular} &
      \begin{tabular}[c]{@{}c@{}}RGBD-\\ based\end{tabular} &
       &
      \begin{tabular}[c]{@{}c@{}}Proposal \\ generation\end{tabular} &
      \begin{tabular}[c]{@{}c@{}}Mask \\ labeling\end{tabular} &
       &
       &
       &
       &
       \\
       \midrule
       \midrule
    \rule{0pt}{2ex}OVIR-3D       &        & $\checkmark$ & $\checkmark$ & $\checkmark$ &        & $\checkmark$ & 11.1 & 20.5 & 27.5 & 52.74  \\
    OpenMask3D    & $\checkmark$ &        &        &        & $\checkmark$ & $\checkmark$ & 13.1 & 18.4 & 24.2 & 547.32 \\
    Open3DIS      & $\checkmark$ & $\checkmark$ & $\checkmark$ & $\checkmark$ &        & $\checkmark$ & 18.5 & 24.5 & 28.2 & 187.97 \\
    \midrule
    \rule{0pt}{2ex}Open-YOLO3D   & $\checkmark$ &        & $\checkmark$ &        &        &        & 23.7 & 28.6 & 34.8 & 16.6   \\
    Ours          & $\checkmark$ & $\checkmark$ & $\checkmark$ &        &        &        &   24.0 (+0.3)   &   31.8 (+3.2)   &   37.4 (+2.6)   & 43.7  \\ 
    \bottomrule
    \end{tabular}%
    }
\caption{Performance and runtime comparison on Replica dataset.}
\label{tab:replica}       
\end{table*}
\paragraph{ScanNet200.} 
We summarize the quantitative evaluation on ScanNet200 in Tab.~\ref{tab:scannet200}. Open-YOLO 3D and our method are significantly faster than other approaches that rely on SAM \cite{sam} for 2D segmentation and CLIP \cite{clip} for prompt label assignment. Our method improves mAP for tail classes compared to Open-YOLO 3D, confirming that 3D models struggle with rare objects and that leveraging 2D foundation models can alleviate this limitation due to the rich world knowledge encoded in 2D models. Note that our approach uses the same pretrained models as Open-YOLO 3D, yet it can effectively leverage the recognition power of 2D foundation models to detect rare classes. This highlights a promising direction for future work: developing strategies that leverage 2D models to enhance recognition of rare objects in real-world applications, while preserving computational efficiency.

\paragraph{Replica.} 
Tab.~\ref{tab:replica} presents results on the Replica dataset. Both Open-YOLO 3D and our method are faster than other approaches relying on SAM and CLIP, while achieving higher overall performance. Compared to Open-YOLO 3D, our method consistently improves mAP across different IoU thresholds, thanks to novel objects discovered via our box-guided proposal generation process. Larger gains are observed at IoU thresholds of 50\% and 25\%, as our pipeline relies solely on superpoints to form novel instance masks. This process can introduce noise and result in imperfect masks, which are penalized at higher IoU thresholds.
\subsection{Qualitative Results}
\begin{figure}[t]
    \centering
    \includegraphics[width=\linewidth]{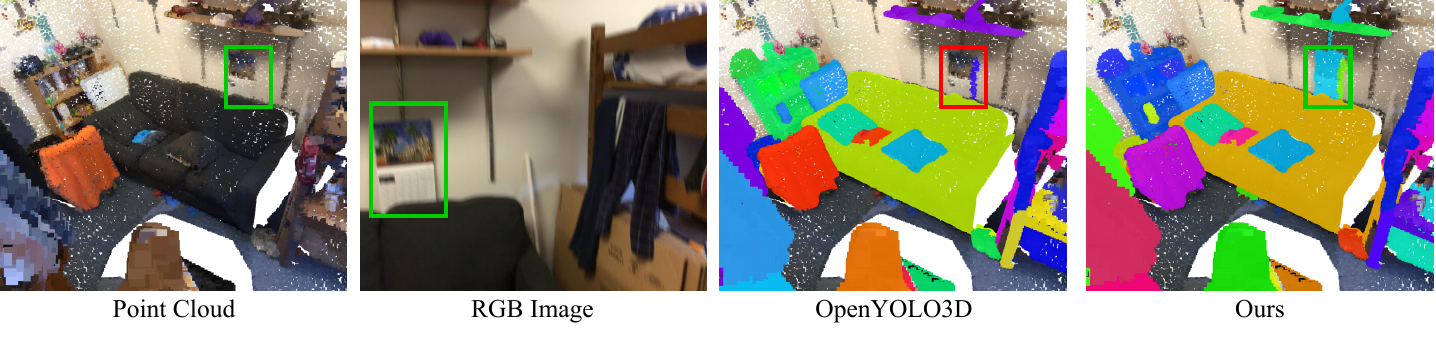}
    \caption{Visualized predictions of Open-YOLO 3D and our method on scene \texttt{scene0353\_00} in ScanNet200 dataset. Open-YOLO 3D completely relies on the pretrained 3D network for instance proposals, unable to retrieve the ``calendar'' instance from the scene while our method can leverage the 2D detector to form extra masks to cover low-frequency classes.}
    \label{fig:visualized_result}
\end{figure}
We visualize the results of Open-YOLO 3D and our approach in Fig.~\ref{fig:visualized_result}. Due to limited 3D training data, 3D instance segmenters often fail to detect objects from low-frequency classes. As a result, methods that rely solely on them, such as Open-YOLO 3D, struggle to generalize to novel categories. In contrast, our method leverages a pretrained 2D object detector to guide the retrieval of rare objects, such as the ``calendar'', which appear infrequently in the 3D training data. This capability makes our approach better suited for real-world applications, where robots or agents must interact with 3D environments that are not fully represented in existing 3D data.
\section{Conclusion}
\label{sec:conclusion}
In this paper, we present a novel approach for retrieving objects in 3D point clouds from language queries via open-vocabulary 3D instance segmentation. Central to our method is box-guided proposal generation, which leverages predictions from an open-world 2D detector to merge 3D superpoints into instance masks for rare objects relevant to the input query. Experiments on two widely used datasets demonstrate that our method outperforms point-based-only approaches, particularly for object categories underrepresented in limited 3D training data.

\paragraph{Limitations and Future Work.}
Although our method does not rely on SAM or CLIP, it still incurs more computation compared to Open-YOLO 3D. The main bottleneck lies in uplifting 2D predictions to 3D, which is relatively slow when obtaining the oriented 3D bounding boxes using Open3D library. Developing a more efficient GPU-based implementation is a promising direction to reduce runtime. While our novel instance masks can capture rare objects, their quality can be further improved. Future work could explore strategies to efficiently refine only the final mask candidates, potentially leveraging SAM, to enhance mask accuracy without sacrificing efficiency. Additionally, leveraging open-world 3D object classifiers such as OpenShape \cite{openshape} or DuoMamba \cite{duomamba} offers a promising alternative for efficiently classifying 3D instances.

\section{Acknowledgments}
This research was supported by the Australian Government through the Australian Research Council's Discovery Projects funding scheme (project DP240101926). Professor Ajmal Mian is the recipient of an ARC  Future Fellowship Award (project FT210100268) funded by the Australian Government.

\bibliography{aaai2026}

\end{document}